\documentclass[letterpaper, 10 pt, conference]{ieeeconf}  

\IEEEoverridecommandlockouts                              
\newcommand{\figref}[1]{Figure~\ref{fig:#1}}

\newcommand{\eqqref}[1]{Equation~\ref{eq:#1}}

\usepackage{caption}

\usepackage{caption}
\usepackage{subcaption}
\usepackage{balance}
\usepackage{booktabs}
\usepackage[noend]{algpseudocode}

\usepackage{array}
\usepackage{graphics} 
\usepackage{epsfig} 
\usepackage{mathptmx} 
\usepackage{times} 
\usepackage{amsmath} 
\usepackage{amssymb}  

\title{\LARGE \bf
{Resource-Efficient Wearable Computing for Real-Time Reconfigurable Machine Learning: A Cascading Binary Classification}
}

\author{Mahdi Pedram, Seyed Ali Rokni, Marjan Nourollahi,
Houman Homayoun, Hassan Ghasemzadeh 
	\thanks{M. Pedram, A. Rokni, M. Nourollahi and H. Ghasemzadeh are with the School of Electrical Engineering and Computer Science, Washington State University (WSU), Pullman, WA, 99164-2752, USA e-mail: \{mahdi.pedram, s.roknidezfooli, m.nourollahidarabad, hassan.ghasemzadeh\} @wsu.edu.}
\thanks{H. Homayoun is with the Department of Electrical and Computer Engineering at George Mason University, Fairfax, VA, 22030-4444, USA e-mail: hhomayou@gmu.edu.}
}

\begin{document}

\maketitle
\thispagestyle{empty}
\pagestyle{empty}

\begin{abstract}

Advances in embedded systems have enabled integration of many lightweight sensory devices within our daily life. In particular, this trend has given rise to continuous expansion of wearable sensors in a broad range of applications from health and fitness monitoring to social networking and military surveillance. Wearables leverage machine learning techniques to profile behavioral routine of their end-users through activity recognition algorithms. Current research assumes that such machine learning algorithms are trained offline. In reality, however, wearables demand continuous reconfiguration of their computational algorithms due to their highly dynamic operation. Developing a personalized and adaptive machine learning model requires real-time reconfiguration of the model. Due to stringent computation and memory constraints of these embedded sensors, the training/re-training of the computational algorithms need to be memory- and computation-efficient. In this paper, we propose a framework, based on the notion of online learning, for real-time and on-device machine learning training. We propose to transform the activity recognition problem from a multi-class classification problem to a hierarchical model of binary decisions using cascading online binary classifiers. Our results, based on Pegasos online learning, demonstrate that the proposed approach achieves 97\% accuracy in detecting activities of varying intensities using a limited memory while power usages of the system is reduced by more than 40\%.

\end{abstract}


\section{Introduction}
\vspace{-2mm}
Wearables have emerged as a promising technology for a large number of application domains such as environmental and medical monitoring \cite{rokni2016smart}, assistance of industry workers \cite{workers}, automatic security surveillance \cite{surveillance}, and home automation \cite{cooksmart}. The utility of wearables in daily life settings is increasing as these systems can be utilized almost anywhere and at anytime. Today, most mobile phones have many embedded sensors such as accelerometers, gyroscopes and magnetometers. Furthermore, the utility of other wearable devices such as smart wrist-bands and watches, necklaces, sports shoes, insoles, and sensors embedded in clothing is growing. In most applications, a sensor node is expected to acquire physical measurements, use embedded software such as machine learning and signal processing algorithms to perform local processing, and store or communicate the results to a gateway or through the Internet \cite{hassan}.

Computational algorithms offer core intelligence of these systems by allowing for continuous and real-time extraction of clinically important information from sensor data. However, many of these algorithms do not consider memory and computation limitations of these lightweight embedded systems for continuous reconfiguration of the underlying model. Designing a personalized algorithm that achieves an acceptable accuracy while satisfying resource constraints of the system is challenging. 

Training activity recognition models on embedded devices can be computationally intensive and requires huge amount of labeled training data. However, embedded systems have a small memory storage and may not be able to store all required labeled data. Furthermore, they are often battery-operated, which makes performing a computationally intensive process on these devices highly challenging. In this paper, we investigate how powerful activity recognition models can be developed using a small storage capacity and limited computation power. Specifically, we investigate the effectiveness of online learning algorithms, which are computationally-simple and require a small memory storage, for detecting physical activities. We propose to transform the multi-class classification problem into a cascading binary classification problem that achieves a high classification accuracy while consuming nominal energy and memory storage.

\section{Problem Statement}
\vspace{-2mm}
A sensing device usually has several sensors for capturing different physical states of the user (e.g., body acceleration, velocity), an embedded software module to perform limited signal processing, machine learning and information extraction, and a radio for data transmission. Conventional approach for dealing with computation and memory limitation of the embedded sensor node is to use a base-station, which is a more powerful unit and responsible for implementing the required power-hungry computational algorithms. However, this model has several shortcomings. First, it prevents the embedded devices from operating independently and thus introduces a single point of failure to the system. Second, the system and sensing nodes cannot provide real-time outcome. Additionally, transmission of the collected data to the base station is still a power-hungry process. Finally, the simple security bridge may introduce privacy concerns. These problems motivate development of a per-node decision making module that is resource-efficient.

An observation $X_i$, sensor readings over a signal segment, made by a wearable sensor at time `$i$' can be represented as a $K$-dimensional feature vector, $X_i$ = \{$d_{i1}$, $d_{i2}$, $\dots$, $d_{iK}$\}. Each feature is computed from a given time window and a marginal probability distribution over all possible feature values. The activity recognition task is composed of a label space $A=\{a_1, a_2, ..., a_m\}$ consisting of the set of labels for activities of interest, and a conditional probability distribution $P(A|X_i)$ which is the probability of inferring label $a_j \in A$ given an observed instance $X_i$. 

Let, for a sensing device $s$, $f$ be the sampling frequency. For each observation $x_i$, the amortized computation complexity of feature extraction, $Compute(x_i)$, is defined as the normalized number of required instructions to extract a feature vector from a signal segment. Furthermore, assume that the learning algorithm requires $M$ blocks of memory where each block is the memory required to store one segment of an activity observation. Assuming $k$ as the maximum available memory in our embedded sensor node, an obvious memory constraints is as follows.

\vspace{-2mm}

\begin{equation}
M \leq k 
\end{equation}

\vspace{-2mm}

Furthermore, we define the minimum acceptable accuracy by introducing a bound on the classifier errors. In other words, for a set of observations $\mathcal{D}$, we have:
\vspace{-2mm}

\begin{equation}
\sum_{x_i \in \mathcal{D}} (y_i - f(x_i)) \leq \epsilon
\end{equation}

\vspace{-2mm}

\noindent where $\epsilon$ denotes the maximum bearable error of the classification. In addition, the power consumption of real-time classification could mainly be divided into the power consumption due to sensing and computation: 

\vspace{-2mm}

\begin{equation}
\label{eq:power}
 f \times(Sense(x) + Compute(x))\leq \theta
\end{equation}

\vspace{-2mm}

\noindent where $\theta$ represents maximum allowable energy consumption. Assuming there exist a possible solution that satisfies these three constraints, we can keep two of the constraints and minimize the third one. For example, accuracy optimization can be defined as accurately labeling observations made by the embedded sensing device such that the mislabeling error is minimized while satisfying the power and memory constraints. Similartly, we can define power and memory optimization problems. However, in this study, we investigate methods of reducing the power and memory consumption while maintaining a high level of classification accuracy.

\section{Cascading Binary Classifier}
\vspace{-2mm}
The problem of activity recognition, as a multi-class classification, requires careful feature design to distinguish all types of activities of interests. However, the computational complexity of these methods increases as the number of classes increases. Having a single multi-class classifier to develop a model to distinguish a class from many other classes is challenging and does not necessarily result in a strong classifier. Alternatively, following Divide-and-Conquer paradigm, decomposing the problem into multiple hierarchical     binary classifier, we can save memory and computation without sacrificing on the classification accuracy.

\subsection{Sensing Efficiency}
\vspace{-1mm}
Attributes of human activities vary from one activity to another. For instance, the activity speed or intensity varies from one activity to another. To develop a single multi-class classification model, we need to capture all required changes of input signal by setting the sampling frequency high enough according the fastest or most intense activity in the set of activities of interest. For many low intensity activities with slow changes in input signal, however, a low sampling frequency is sufficient. In addition, a multi-class classification problem requires a complex set of features to separate all activities of interest while both feature extraction and sensing need computation and consume power. 

In this study, we design a cascading method to efficiently adapt sampling frequency and extracted feature complexity. Similar to Metabolic Equivalent
of Task (MET) studies, we categorize all activities of interest into three categories: low-intensity, medium-intensity, and high-intensity \cite{alinia2016reliable}. For low-intensity activities, a low sampling frequency is sufficient to capture the required information for detecting the activity. Similarly, for high-intensity activities, we need higher sampling frequency to precisely detect such activities. 
 
\begin{figure*}[tbh]
\hspace{-10mm}\centering
\includegraphics[scale=.64]{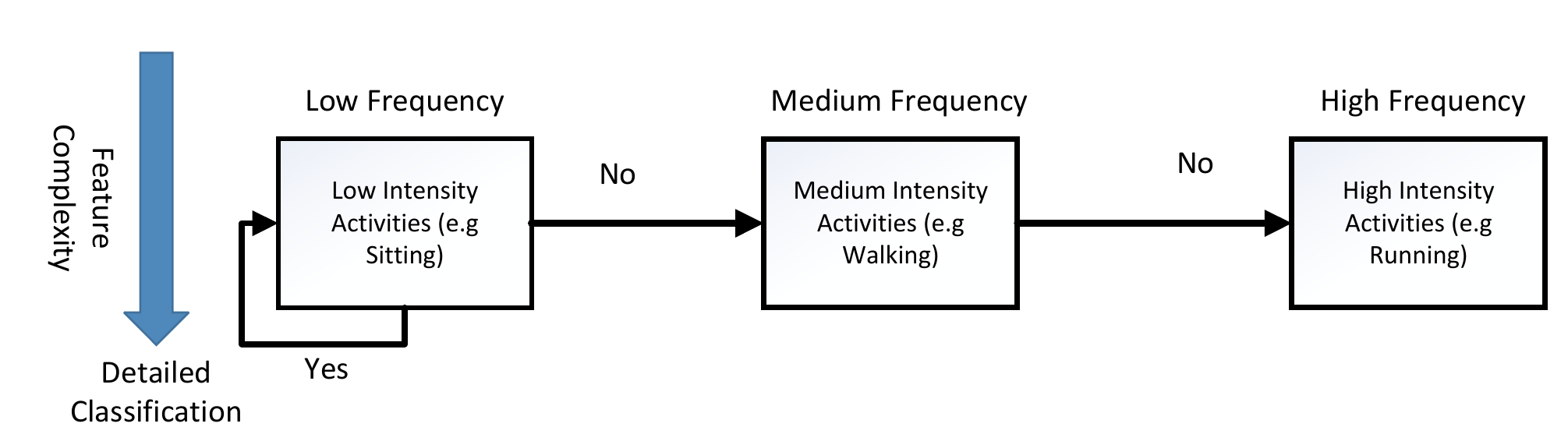}
\vspace{-2mm}
\caption{Cascading binary classification model.}
\label{fig:solution}
\vspace{-5mm}
\end{figure*}

As shown in in \figref{solution}, we first categorize activities of interest into three clusters of (low, medium, and high) intensity. Next, we start with low frequency and computationally simple features such as amplitude of the signal. Using this approach we develop a binary classifier to separate low-intensity activities from the remaining activities. If the classifier identifies the current activity as low-intensity, the sampling rate remains low. Further detailed classification for separating low-intensity activities such as `sitting' versus `standing' performs with higher representative features but in low frequency. On the other hand, if the classifier detects the current activity as a higher-intensity activity, we increase the sampling rate to medium level and use the second level of features and the classifier for separating medium-intensity activities from high-intensity activities. We also use three level of feature vectors from simple to complex according to their computing complexity. First, we detect the intensity class of a given activity signal. Moving to the right in \figref{solution}, we understand that the activity is in higher intensity class and therefore we increase the sampling frequency and examine for more intense activities. Knowing the intensity class of the activity, without changing the sampling frequency, we move from up to down in \figref{solution}, use more complex features and distinguish classification labels of activities in the same category of the activity intensity. 

The power consumption of the cascading binary classifier framework is given by

\vspace{-6mm}
\begin{multline}\label{eq:power2}
(T_{l} ~f_{l} (Sense(x) + Compute_{l} (x) + Compute_{m}(x))\\ +
(T_{m}~ f_{m} (Sense(x) + Compute_{m}(x) + Compute_{h}(x)) \\+
(T_{h} ~f_{h} (Sense(x) + Compute_{h}(x))
\end{multline}

\noindent where $T_{l}$, $f_{l}$ and $Compute_{l}$ are accumulated duration low intensity activity, frequency for detecting low intensity vs. non-low intensity activities and computation of first level of feature extraction. We add $Compute_m$ for the low-intensity activities because after knowing that the activity is of low-intensity, the classifier uses a second level of features to distinguish different low-intensity activities and infers one single activity as the classification result. The process is similar for activities of higher intensities (i.e., medium, high).

\subsection{Memory Efficiency}
As previously discussed, embedded systems often have a small memory storage. To reduce the amount of memory usage, we use online learning methods which are memory efficient. Particularly, each block in \figref{solution} is a Pegasos classifier \cite{shalev2007pegasos} which runs in mini-batch mode. The size of mini-batch is determined by $k$, the amount of available memory. 

Pegasos is a simple and effective stochastic sub-gradient descent algorithm for solving the optimization problem cast by Support Vector Machines (SVM). The algorithm utilizes $k$ examples during each iteration, where $k$ is the given memory constraint. The original Pegasos chooses a subset $A_t \subset [m] = \{1, \dots ,m\}$, $|A_t| = k$, of $k$ examples uniformly at random among all such subsets during each iteration of the learning. However, in our problem, we assume that we have only $k$ instances of observation at each time. Therefore, upon any new observation, we update the list of $k$ observation by eliminating the least recent observation and adding the newest one. Pegasos considers the following approximate objective function.

\begin{equation}
f(w;A_t) = \lambda/2 \lVert w\rVert^2 + 1/k \sum_{i \in A_t} l(w;(x_i, y_i)).
\end{equation} 

It then considers the sub-gradient of the approximate objective given by

\begin{equation}
\bigtriangledown_t = \lambda w_t - 1/k \sum_{i \in A_t} 1 [y_i \langle w_t, x_i \rangle < 1] y_i x_i
\end{equation}

\noindent and updates using $w_{t+1}\leftarrow w_t - \eta_t \bigtriangledown_t$ based on the same predetermined step size $\eta_t = 1/ (\lambda_t)$. Therefore, the only difference with the original Pegasos is that we do not have access to all dataset in advance for the purpose of sampling. On streaming  data, we only have access to $k$ recent observations  at any time and perform that sub-gradient based on available observations.

\section{Validation}
\vspace{-2mm}
\subsection{Experimental Setup}
\vspace{-1mm}
Subjects performed the activities at Bilkent University Sports Hall while wearing $5$ motion sensor nodes on different body locations.Each sensor node is a Xsens MTx inertial sensor unit which has a $3$-axis accelerometer, $3$-axis gyroscope and $3$-axis magnetometer. The sampling frequency was set to $25$Hz and the 5-min signals are divided into 5-sec segments so that 480(=60x8) signal segments are obtained for each activity. 



We selected 7 activities with different intensity levels as a proof of concept in this paper. Sitting and standing are low intensity activities. Walking in a parking lot and walking on a treadmill with a speed of 4 km/h in flat are  medium intensity activities. Running, exercising, and jumping are high intensity activities.



From each 5-sec segment of the individual sensor streams, we extracted $9$ statistical features. Potentially, there are many different features that can be extracted from human activity signals. However, we use amplitude of the signal(AMP), median of the signal(MED), mean of the signal (MNVALUE), maximum of the signal (MAX), minimum of the signal (MIN), peak to peak amplitude (P2P), standard deviation of the signal(STD), root mean square power(RMS), and stand to end value(S2E). These features aimed to capture both shape and amplitude of the signals. For instance, features such as amplitude of the signal and mean of the signal are useful to capture amplitude of the signal while standard deviation of the signal and start to end value attributes attempt to capture the structure of the signal. We divided these features into three levels of complexity where AMP was used in the first level, AMP, MNVALUE, and STD were used in the second level, and all nine features were used in the third level.

%
%
%
%

\begin{figure}[tbh]
\centering
\includegraphics[width=0.48
\textwidth]{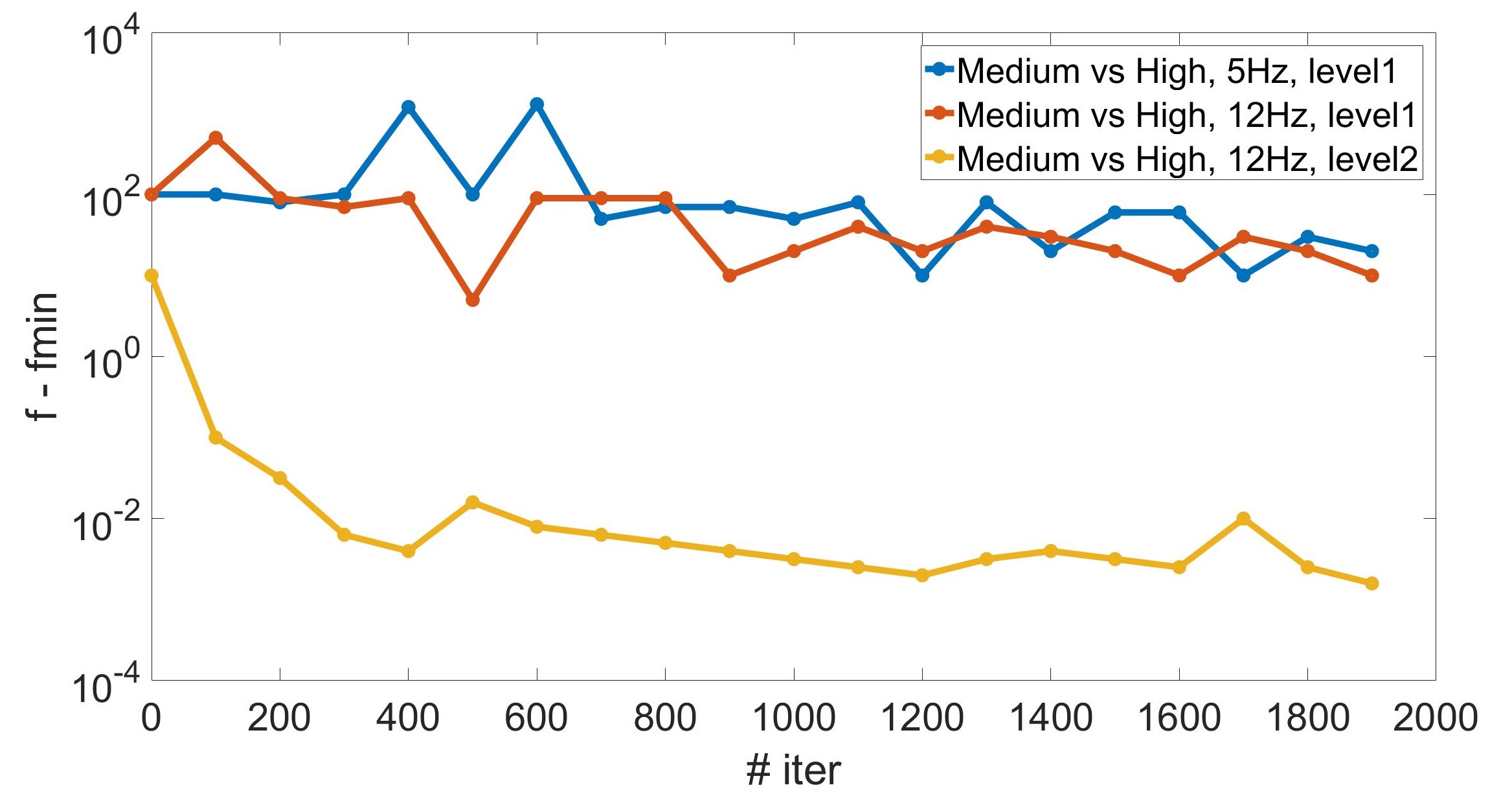}
\caption{Convergence of Medium vs. High in different frequency and feature level}
\label{fig:accelerometer_data_during_drinking}
\vspace{-7mm}
\end{figure}

\subsection{Results}
\vspace{-2mm}
We used Pegasos algorithm for dealing with memory constraints. Particularly, in this experiments we set the value of $k$ to $10$. At the first level, we aimed to detect the current activity as low-intensity or not-low-intensity activity. Therefore, we developed a binary classifier using Pegasos while feeding with low = $1$ and not-low = $-1$ as binary class labels. Because our goal was to do sensing in low frequency and the dataset frequency is 25Hz, we down-sampled the signals to 5Hz. As discussed in previous sections, we use AMP as the only feature in this level. Only using a 5Hz sampling rate and level 1 features we achieved $100$\% accuracy to distinguish low-intensity activities from others. However, only using this single feature is not enough for classification among low-intensity activities. For example, for separating \textit{sitting} from \textit{standing} using level 1 features we obtain an accuracy of 51\% which is close to random guess. By using more complex features (level 2 features) we achieve $100$\% accuracy for detecting \textit{sitting} vs. \textit{standing} at a frequency of $5$Hz.

As shown in \figref{accelerometer_data_during_drinking}, when low versus not-low classifier detects the current activity as not-low, we need to detect if the intensity of activity is medium or high. As previously discussed, in this level we cannot distinguish activities using sparse samples of the low-intensity level. With only low sampling rate and level 1 features the objective function highly fluctuates. Blue line in \figref{accelerometer_data_during_drinking} shows binary classifier medium versus high-intensity activities does not converge using only low frequency sampling. On the other hand, by only increasing the frequency and maintaining the level 1 feature, it reaches an accuracy of $59$\% which is close to random guess because after filtering low-intensity activities $60$\% of the remaining activities were of high-intensity. After increasing the sampling and using level 2 features, we were able to perfectly classify instances of medium activities from instances of high-intensity activities.


When the intensity class of an activity is detected, we fix the sampling rate accordingly, and use more complex feature to distinguish different activities in the same intensity class. For example, to classify \textit{walking in parking lot} from \textit{walking on treadmill}, we used second level frequency, 12 Hz in this study. At this frequency with level 2 features, we gain an accuracy of $95\%$.
 When we keep the frequency at 12Hz and do feature extraction at level 3, we achieve a better convergence and the classifier shows an accuracy of $97$\% at the cost of higher number of iterations. Similarly, when the intensity class of activity is detected as high, we need higher sampling rate to classify individual activities within high intensity activities. At 25Hz sampling rate and using level 3 features, we can classify each high-intensity activity from other high-intensity activities with an average accuracy of $97$\%. 

\subsection{Power and Memory Analysis}
\vspace{-1mm}
According to \eqqref{power}, the corresponding power consumption of activity recognition is proportional to its sensing and computation overhead. In this paper we assume each feature could be computed with $O(n)$ instructions where $n$ is the number of samples in the signal segment. Therefore, all the features are assumed to have equal computation overhead. In the case of traditional multi-class classifier, the power consumption of the activity recognition algorithm can be approximated by \eqqref{power} where we use $f$ = $25Hz$ and level 3 features for all activities. On the other hand, the power consumption of the cascading binary classifier could be approximated using \eqqref{power2}. Assuming the likelihood of occurring each activity is the same as other activities, our proposed method leads to $44$\% power savings in sensing and $42$\% power savings in feature computation.

In terms of memory usage, the traditional multi-class classifier requires all the training data of all activities to be stored. Therefore, the required storage increases as the size of the training data increases. However, using Pegasos, the memory usage remains constant regardless of the size of training data. For example, in our experiments, we need a small memory for storing only $10$ segments of the data.

\section{CONCLUSIONS}
\vspace{-1mm}
To develop a personalized machine learning model for wearables, these models need to be retrained upon any changes in configuration of the system. However, the available memory and computing power on these lightweight devices is quite limited. This limitation warrants development of personalized and adaptive machine learning algorithms to be efficient in terms of computation costs and memory usage. In this paper, we designed a resource-efficient framework for classification of human activities using hierarchical classifier design and online learning models. We showed that using cascading online binary classifiers, the system achieves a reasonable classification accuracy with small memory usage while achieving over $40$\% saving in power consumption.

\addtolength{\textheight}{-12cm}   





\section*{ACKNOWLEDGMENT}
\vspace{-1mm}
The authors would like to thank  Dr. Janardhan Rao (Jana) Doppa of Washington State University and Dr. Sourabh Ravindran of Samsung Research America for their valuable input and technical discussions. This work was supported in part by the United States Department of Education, under Graduate Assistance in Areas of National Need (GAANN) Grant P200A150115, and the United States National Science Foundation, under grant CNS-1750679. Any opinions, findings, conclusions, or recommendations expressed in this material are those of the authors and do not necessarily reflect the views of the funding organizations.


\bibliographystyle{IEEEtran} 
\bibliography{myrefs}

\end{document}